\documentclass[conference, a4paper]{IEEEtran}
\IEEEoverridecommandlockouts
\usepackage{cite}
\usepackage{amsmath,amssymb,amsfonts,bm}
\usepackage{algorithmic}
\usepackage{graphicx}
\usepackage{xcolor, balance}
\usepackage{amsfonts,amsthm}
\usepackage{listings,algorithmic,algorithm}
\usepackage{epsfig}
\usepackage{xspace}

\theoremstyle{definition}

\theoremstyle{remark}

\usepackage{color}  

\definecolor{darkgreen}{rgb}{0., 0.4, 0.}

\newcommand{\cp}[1]{\mathcal{#1}}
\newcommand{\calN}{\cp{N}}

\newcommand{\bF}{\boldsymbol{F}}
\newcommand{\bG}{\boldsymbol{G}}

\newcommand{\bI}{\boldsymbol{I}}

\newcommand{\bM}{\boldsymbol{M}}

\newcommand{\bP}{\boldsymbol{P}}
\newcommand{\bQ}{\boldsymbol{Q}}
\newcommand{\bR}{\boldsymbol{R}}
\newcommand{\bS}{\boldsymbol{S}}

\newcommand{\bp}{\boldsymbol{p}}

\newcommand{\br}{\boldsymbol{r}}
\newcommand{\by}{\boldsymbol{y}}

\newcommand{\bx}{\boldsymbol{x}}

\newcommand{\bbR}{\mathbb{R}}

\newcommand{\bmu}{\boldsymbol{\mu}}
\newcommand{\btheta}{\boldsymbol{\theta}}

\newcommand{\bSigma}{\boldsymbol{\Sigma}}

\usepackage{nccmath}    
\useshortskip 

\newcommand{\cb}[1]{\boldsymbol{#1}}

\newcommand\numberthis{\addtocounter{equation}{1}\tag{\theequation}}

\begin{document}
\allowdisplaybreaks

\title{Hybrid Neural Network Augmented Physics-based Models for Nonlinear Filtering
\thanks{This work has been partially supported by the NSF under Awards CNS-1815349 and
ECCS-1845833, NSF-1935337. 
}
\thanks{{$^\ast$} Indicates shared first authorship.}}

\author{
\IEEEauthorblockA{Tales Imbiriba$^{\dagger\ast}$, Ahmet Demirkaya$^{\dagger\ast}$, Jind{\v{r}}ich Dun\'{i}k$^\ddagger$, Ond{\v{r}}ej Straka$^\ddagger$, Deniz Erdo{\u{g}}mu\c{s}$^\dagger$, and Pau Closas$^\dagger$}
\IEEEauthorblockA{$\dagger$ Electrical \& Computer Engineering Department\\
Northeastern University \\
Boston, MA, USA}
Emails: \{talesim, demirkaya, erdogmus, closas\}@ece.neu.edu
\IEEEauthorblockA{$\ddagger$ Department of Cybernetics\\
University of West Bohemia \\
Pilsen, Czech Republic}
Emails: \{dunikj, straka30\}@kky.zcu.cz
}

\maketitle

\begin{abstract}
In this paper we present a hybrid neural network augmented physics-based modeling (APBM) framework for Bayesian nonlinear latent space estimation. The proposed APBM strategy allows for model adaptation when new operation conditions come into play or the physics-based model is insufficient (or incomplete) to properly describe the latent phenomenon. One advantage of the APBMs and our estimation procedure is the capability of maintaining the physical interpretability of estimated states. Furthermore, we propose a constraint filtering approach to control the neural network contributions to the overall model. We also exploit assumed density filtering techniques and cubature integration rules to present a flexible estimation strategy that can easily deal with nonlinear models and high-dimensional latent spaces. Finally, we demonstrate the efficacy of our methodology by leveraging a target tracking scenario with nonlinear and incomplete measurement and acceleration models, respectively. 
\end{abstract}

\begin{IEEEkeywords}
Nonlinear filtering; Target tracking; Hybrid Neural Network; Physics-based Neural Models; Gaussian filtering. 
\end{IEEEkeywords}

\section{Introduction}

Complex nonlinear dynamic models can be found in numerous applications such as describing biological systems, weather prediction, fluid dynamics, and target tracking, to name but a few. In many such applications the ``true'' underlying model can be very complex and context-dependent. For instance, this is the case for sunlight interaction with materials in the Earth surface~\cite{imbiriba2016nonparametric, borsoi2020kalman}, disease spread~\cite{arora2020prediction}, indoor positioning \cite{Dardari15}, navigation \cite{dunik2020state}, or predicting the trajectory of a target~\cite{imbiriba_gppf_2020}. In these examples, the model governing the evolution of such signals is often either very complex or unknown (in the case of target tracking) leading to more complex optimization strategies and/or other sources of information to be able to provide accurate predictions. 

In this context machine learning (ML) algorithms become appealing. This is the case especially when accurate physical-based models are too complex or unknown. One drawback of purely data-driven ML strategies relates to the interpretability and physical meaning of estimated quantities especially when one aims at recovering latent states~\cite{Li_DeepAECUnmixing2021, imbiribaPlans2020indoor}.

In this contribution, we are especially interested in leveraging ML strategies to improve nonlinear dynamical models. In this context, ML approaches can be classified into hybrid~\cite{psichogios1992hybrid,retina_embc_2021}, where data driven models are used in combination with physics-based models, or purely data-driven following an end-to-end learning philosophy~\cite{haykin2004kalman}. 
%
%
%
The hybrid approaches focus on providing corrections to estimates. The algorithms~\cite{chin1994a} and~\cite{vaidehi1999a} use back-propagation NNs to correct position estimates.The algorithm proposed in~\cite{owen2003a} predicts nonlinear velocity and acceleration corrections to linear predicted states by a recurrent NN and the algorithm~\cite{shaukat2021a} augments the error-state Kalman filter (KF) by an RBF NN to compensate for the lack of KF performance. The algorithms use NNs that are unaware of the system model.

In addition to providing estimate corrections, several hybrid approaches directly estimate the state. In~\cite{gao2019a} two algorithms based on deep long short-term memory (LSTM) NNs were proposed, which provide estimates either in two steps (time-update and measurement-update) or in a single step. 
Both algorithms are unaware of the system model and the learning is based on estimated quality optimization. The algorithm~\cite{jung2020a} uses the NN in the prediction step of the KF to provide not only the estimate but also the associate covariance matrix. In this case, the learning optimizes the negative log-likelihood of multivariate normal distribution. An alternative approach was proposed in~\cite{zhao2019a}, where the NN was used to learn the system model parameters such as state transition and measurement matrices, and associated noise covariance matrices under the framework of Bayesian estimation. Recently, deep Kalman filters strategies were proposed in a smoother-like approach~\cite{krishnan2015deep} and efficient KF implementations where the Kalman gain is approximated by NNs~\cite{revach2022kalmannet}. In both works however, the hybrid APBM model component were not studied. 

The data-based approaches use plain data to learn the mapping from observations to the states to avoid complications of the hybrid methods.
In~\cite{thormann2017a} a large amount of data was simulated to be able to achieve end-to-end learning while in~\cite{zhai2019a} all components of the process, i.e., data generator, sliding window, centralization strategy, and the learner, have been proposed.



In this paper we are concerned about nonlinear dynamical models of the type:
\begin{align}
    \dot{\bx}_t &= f(\bx_t) + \cb{q}_t \label{eq:dyn_ct}\\
    \by_t &= h(\bx_t) + \br_t \nonumber
\end{align}
where $\bx_t\in\bbR^{d_x}$ is the state vector, $f$ is the state transition function and $\cb{q}_t$ is an arbitrary zero-mean noise term independent of $\bx_t$. More specifically, we are interested in models where the transition function $f$ is not fully representative of the true state dynamics due to either simplistic physics-based models not being capable to explain time-varying dynamic scenarios or other unmodeled factors. In such a scenario, we propose a hybrid neural network augmentation approach capable of augmenting physics-based models. To cope with time-varying dynamics we use a continuous learning strategy consisting of augmenting the states with model parameters leading to recursive Bayesian estimation (RBE) approaches~\cite{wu2019wifi}. One of the challenges of incorporating neural network parameters as states in an RBE approach is related to the computational complexity inherent to high-dimensional state spaces especially when nonlinear or non-Gaussian models are in play. To partially mitigate this issue, we leverage Cubature integration strategies in association with Gaussian assumptions, i.e., the Cubature Kalman filter~\cite{arasaratnam2009cubature}. We also propose a strategy to control the neural network contribution to the overall dynamic model.

This work is organized as follows. In Section~\ref{sec:HNNPBM} we present our NN augmentation strategy. In Section~\ref{sec:Control} we present an augmented likelihood mechanism to control the NN's contribution. In Section~\ref{sec:Cubature} we discuss the Cubature integration and relationship with different filtering approaches. Experiments are presented in Section~\ref{sec:exp} and final remarks discussed in Section~\ref{sec:conc}.



\section{Hybrid Neural Network Physics-based models}
\label{sec:HNNPBM}

In this section, we aim at augmenting the ODE model in~\eqref{eq:dyn_ct} using neural networks. The model mismatch might occur due to missing ODE components or inaccurate models that do not match the governing physics of the phenomenon. In this contribution, we aim at the latter. That is, we assume that the number of states is known and somehow represented by the physical model $f$.
The discretization of~\eqref{eq:dyn_ct} is 
\begin{align} 
    \bx_k &= \bx_{k-1} + T_s f(\bx_{k-1}) + T_s\tilde{\cb{q}}_{k-1} \label{eq:dyn_discrete}\\
    \by_k &= h(\bx_k) + \br_k \nonumber
\end{align}
where $T_s$ is the sampling period and $\tilde{\cb{q}}_{k-1}$ is the independent discretized noise term. Thus, we propose to augment the discretized dynamical model in~\eqref{eq:dyn_discrete} as:

\begin{align} \label{eq:discretized_model}
    \bx_k =  g\left(f(\bx_{k-1}), \bx_{k-1}; \cb{\theta}\right) + \cb{q}_{k-1}
\end{align}
where $g(\cdot): \bbR^{d_x}\times \bbR^{d_x} \to \bbR^{d_x}$ is a vector-valued function, modeled as neural network, and parametrized by $\cb{\theta} \in \bbR^{P}$, that we assumed to have incorporated the sampling period $T_s$, and  $\cb{q}_{k-1} = T_s \tilde{\cb{q}}_{k-1}$. The model in Eq.~\eqref{eq:discretized_model} is flexible enough that  allow for both replacing whole ODEs (e.g, $f(\bx_k)= 0$), or fusing arbitrary functions of $\bx_k$ and $f(\bx_k)$. We call models extended in such fashion \emph{augmented physics-based models} (APBMs).

\section{Controlling neural network contributions in augmented models}
\label{sec:Control}

One important point regarding the hybrid dynamical model is the capability of controlling the contribution of the neural augmentation. In this section we propose to introduce a regularization over the NN model parameters $\cb{\theta}_k$ by an augmentation of the likelihood model~\cite{simon2010kalman} as $p(\by_k, \bar{\cb{\theta}}|\bx_k, \cb{\theta}_k)$, where $\bar{\cb{\theta}}$ is the $P$-dimensional vector designed such that $g\left(f(\bx_{k}), \bx_{k}; \cb{\theta}=\bar{\cb{\theta}}\right) =  f(\bx_{k})$.
This allows us to re-write the state-space model in~\eqref{eq:discretized_model} as:
\begin{align}
    \cb{\theta}_k &= \cb{\theta}_{k-1} + \cb{q}_{k-1}^\theta \\
    \bx_k &= g(f(\bx_{k-1}),\bx_{k-1}; \cb{\theta}_{k-1}) + \cb{q}_{k-1}^x \\
    \begin{bmatrix}
    \by_k \\
    \bar{\cb{\theta}}
    \end{bmatrix} & = 
    \begin{bmatrix}
    h(\bx_k) \\
    \cb{\theta}_k
    \end{bmatrix} + 
    \begin{bmatrix}
    \br^{y}_k\\
    \br^{\theta}_k
    \end{bmatrix}
\end{align}
where $\cb{q}_{k-1}^\theta \sim \calN(0, \bQ^\theta)$, $\cb{q}_{k-1}^x \sim \calN(0, \bQ^x)$ model the dynamic model uncertainty, $\br^{y}_k\sim\calN(0, \bR^y)$ is the measurement noise, and $\br^{\theta}_k\sim\calN(0, \frac{1}{\lambda}\bI)$ defines the ball around $\bar{\cb{\theta}}$ of possible solutions for $\cb{\theta}_k$. 
The posterior can then be re-written as
\begin{align}
    p(\bx_k,\cb{\theta}_k &|\by_{1:k}, \bar{\cb{\theta}}) \propto  p(\by_k,\bar{\cb{\theta}}| \bx_k, \cb{\theta}_k) \\
    & \times \int\int  p(\bx_k|\bx_{k-1}, \cb{\theta}_{k-1}) p(\bx_{k-1}| \cb{\theta}_{k-1}, \by_{1:k-1}, \bar{\cb{\theta}}) \nonumber\\
    & \times   p(\cb{\theta}_k| \cb{\theta}_{k-1} ) p(\cb{\theta}_{k-1}| \by_{1:k-1}, \bar{\cb{\theta}})  d\bx_{k-1} d\cb{\theta}_{k-1} \nonumber
\end{align}
In this Gaussian case the recursive Bayesian filtering solution for the above problem is equivalent to the solution to the following problem for the state mean at every time instant $k$:
\begin{align}\label{eq:opt_prob}
    (\hat{\bx}_k, \hat{\cb{\theta}}_k) =& \mathop{\arg\min}_{(\bx, \cb{\theta})}  \|\by_k - h(\bx)\|^2_{\bR^{-1}} + \lambda \|\cb{\theta}- \bar{\cb{\theta}}\|^2  \\
    &+ \|\bx - f(\hat{\bx}_{k-1})\|^2_{[\hat{\bP}_{k|k-1}^x]^{-1}} + \|\cb{\theta} - \hat{\cb{\theta}}_{k-1}\|^2_{[\hat{\bP}_{k|k-1}^\theta]^{-1}} \nonumber
\end{align}
where the terms correspond to data fit, neural network parameter regularization, and the two regularizations resulting from the the dynamical models for the states $\bx$ and $\cb{\theta}$, respectively. 
We call attention to the fact that the parameter $\lambda$ controls the regularization term $\|\cb{\theta}- \bar{\cb{\theta}}\|^2$ and, thus, the contribution of the neural network model. For instance, when $\lambda \to \infty$ the parameter constraint is fully enforced in~\eqref{eq:opt_prob} completely eliminating the neural augmentation contribution to the model. When $\lambda \to 0$ the parameter constraint in~\eqref{eq:opt_prob} is turned-off leading to free neural network model contributions that can, possibly, overpower the physics-based model.

\section{Efficient Gaussian filter model learning}
 \label{sec:Cubature}
In this paper, we train a hybrid ODE and neural network model using recursive Bayesian state estimation. Specifically, when the predictive and observation models are linear and Gaussian, the state posterior recursion integrals 
can be solved analytically leading to the well-known Kalman time and measurement update equations~\cite{sarkka2013bayesian}. When nonlinear models are in play, the required integrals often become intractable and numerical strategies must be sought \cite{closas2015computational}. Alternatives include linearization of nonlinear functions (extended Kalman filters, EKF) or sampling methods such as particle filters~\cite{arulampalam2002tutorial}, unscented Kalman filters (UKF)~\cite{wan2001unscented}, or cubature Kalman filters (CKF)~\cite{arasaratnam2009cubature}, which assume different levels of system simplicity. 
For instance, EKF, UKF, and CKF assume Gaussianity of the measurement and transitional models while handling the integration exploiting this Gaussianity in different ways. While EKF linearizes the models using a first-order Taylor expansion, UKF and CKF use unscented and cubature rules to compute integrals of the form

\vspace{-0.03in}

\begin{align}\label{eq:int_l}
    I(\ell) = \int_\cp{D} \ell(\bx) p(\bx) d\bx
\end{align}

\noindent where $\ell$ is a nonlinear function of $\bx\in\bbR^{d_x}$ and $p(\bx)=\calN(\bmu, \bSigma)$ is a Gaussian PDF with mean $\bmu$ and covariance $\bSigma$, as a weighted sum of function evaluations of a finite number of deterministic points. For the third-degree cubature rule, the integral in~\eqref{eq:int_l} can be approximated as
\begin{align}\label{eq:cubature_int}
    I(\ell) \approx \frac{1}{2d_x} \sum_{j=1}^{2d_x} \ell(\bS^\top\cb{\xi}_j + \bmu) 
\end{align}
where $\cb{\xi}_j = [\cb{1}]_j \sqrt{2d_x/2}$ are deterministic points~\cite{arasaratnam2009cubature}, and $\bS$ is the lower triangular Cholesky decomposition such that $\bSigma = \bS\bS^\top$. It is important to highlight that the cubature rule demands only two points per dimension of $\bx$, i.e., $2d_x$, to evaluate the sum in~\eqref{eq:cubature_int}, making it more suitable when working in high-dimensional state-spaces. Assuming Gaussianity of state posteriors, the CKF can solve the integrals required in the Bayesian recursion as well as the moments (mean and covariance) of the new state posterior~\cite{arasaratnam2009cubature}.

In contrast, particle filters do not assume any particular distribution; instead, they approximate the distribution as a linear combination of Dirac deltas. Thus, moments of propagated particles can be easily computed. One drawback of particle filters is the high number of particles needed to accurately represent distributions. This issue is profoundly aggravated if the state-space dimension is large, making this filtering strategy unfeasible in such scenarios~\cite{imbiriba_gppf_2020}.

\section{Experiments} \label{sec:exp}

In this section, we present two experiments designed to test different forms of model augmentation. In the first, see Section~\ref{subsec:chao}, we consider the Lorenz Attractor~\cite{lorenz_1963} where we replaced a whole ODE with a neural network model. We keep the approach hybrid in the sense that we kept the remaining ODEs fixed. In the second example, see Section~\ref{subsec:target}, we exploit a target tracking example where the data was generated with a model that contains additive constant velocity and sinusoidal components. In this example, we augmented the constant velocity model with a neural network. 

For both examples, we performed Monte Carlo simulations using 100 runs. 
To measure the filtering performance we consider the tracking root mean-squared error (RMSE):
\begin{equation}
    \text{RMSE}_k = \sqrt{\frac{\sum_{r=1}^{N_{MC}}\|\bp^{(r)}_k-\hat{\bp}^{(r)}_k\|^2}{dN_{MC}}}
\end{equation}
with $d$ is the dimension of $\bp_k$, $N_{MC}$ is the number of Monte Carlo runs, and $\hat{\bp}_k$ is the estimated states of interest, e.g., position for the target tracking case and all states for the Lorenz Attractor example.

To analyse the evolution of the weights over time with different $\lambda$, we computed the average variance of parameters as: 
\begin{equation}
    \mathop{\mathbb{E}}\lbrack \text{Var}(\btheta_k) \rbrack = \frac{\sum_{r=1}^{N_{MC}}\text{Var}(\btheta^{(r)}_k)}{N_{MC}} \label{eq:var_formula}
\end{equation}
where $\text{Var}(\btheta^{(r)}_k)$ is the sample variance of the neural network parameters for the $r$-th Monte Carlo run. 

\subsection{Application to chaotic systems} \label{subsec:chao}

In this section we present the Lorenz Attractor \cite{lorenz_1963} as an example chaotic system to test our approach. The Lorenz Attractor consists of three ordinary differential equations (ODEs):
\begin{align*}
\dot{x}_1 & = \sigma (x_2 - x_1)
\numberthis \label{eq:lorenz_x1} \\
\dot{x}_2 & =  x_1 (\rho - x_3) 
\numberthis \label{eq:lorenz_x2} \\
\dot{x}_3 & = x_1 x_2 -  \beta x_3
\numberthis \label{eq:lorenz_x3}
\end{align*}
where $x_i$, $i=1,2,3$, are the system states and $\sigma, \rho, \beta \in \mathbb{R}$ are model parameters. 

The model can be discretized as 
\begin{align}
\bx_k = \bx_{k-1} + T_s f(\bx_{k-1})
\end{align}
where $\bx_k \in \bbR^3$ is the vector of states and $f(\cdot) = [f_1(\cdot), f_2(\cdot), f_3(\cdot)]^\top$ is a vector valued function representing the dynamical model in Eqs.~\eqref{eq:lorenz_x1}--\eqref{eq:lorenz_x3}.
We generate measurements according to the following measurement equation:
\begin{equation}\label{eq:lorenz_meas_eq}
{\mathbf y}_k = \bx_k + {\cb{n}}_k ~,
\end{equation}
where $\cb{n}_k \sim \calN(0, 0.001 \bI)$.
The initial state was set as $\bx_0 = (1, 1, 1)^\top$, and the dynamical model parameters as $\sigma=28$, $\rho=10$, and $\beta=8/3$, and $T_s = 1$s. 

In this first experimental setup, we follow a procedure similar to the one in~\cite{retina_embc_2021} where we replaced an ODE of the system was replaced by a neural network. In the example presented in this section we 
replaced the ODE for $x_1$ in Eq. \eqref{eq:lorenz_x1} with a neural network $\gamma(\cdot)$. In this case, the discretized NN-based version of~\eqref{eq:lorenz_x1} can be written as
\begin{align}
x_{1,k} & = x_{1, k-1} + \gamma(\bx_{k-1}; \cb{\theta})
\label{eq:lorenz_x1_nn} 
\end{align}
leading to:
\begin{equation}
    g(f(\bx_{k-1}), \bx_{k-1}; \cb{\theta}) = 
    \begin{pmatrix}
    x_{1,k-1} + \gamma(\bx_{k-1}; \cb{\theta}) \\
    f_2(x_{1,k-1}, x_{3,k-1})\\
    f_3(\bx_k)
    \end{pmatrix}
\end{equation}
where $f_2$ and $f_3$ represent the functions in the RHS of~\eqref{eq:lorenz_x2} and~\eqref{eq:lorenz_x3}, respectively, and
$T_s$ has been incorporated into the model parameters $\cb{\theta}$. The structure of $\gamma$ was consists of 3 inputs, a hidden layer with 5 hidden units and ReLu activation functions, and a single output neuron with linear activation.





\begin{figure}[htb]
\centering
\includegraphics[trim={0 3cm 0 4cm},clip,width=\columnwidth]{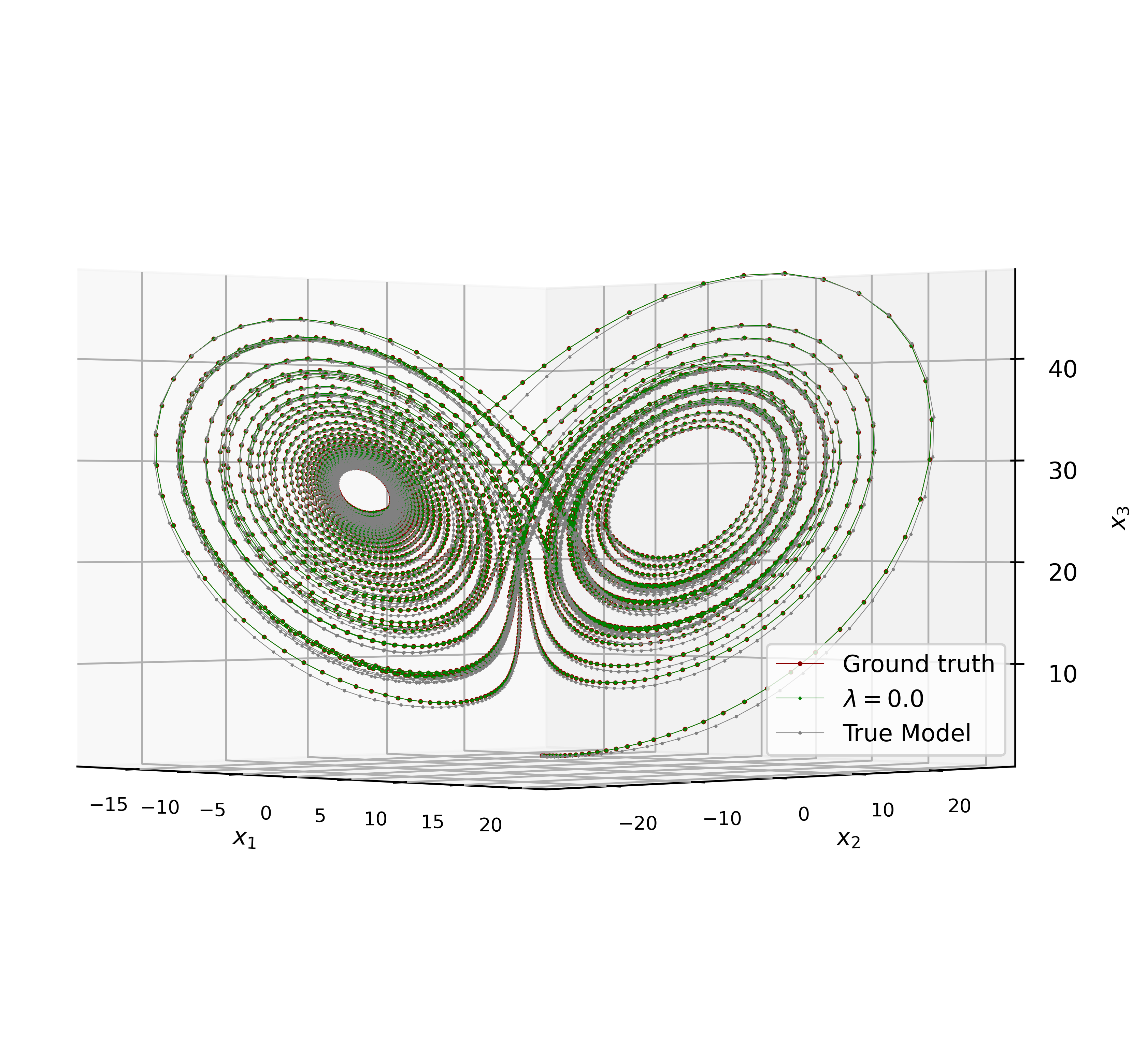}
\caption{Single simulation demonstrating the effectiveness of the APBM while tracking a chaotic system.}
\label{fig:neuron_FFTT}
\end{figure}

\begin{figure}[htb]
\centerline{\includegraphics[trim={0 0.2cm 0 0.1cm},clip,width=\columnwidth]{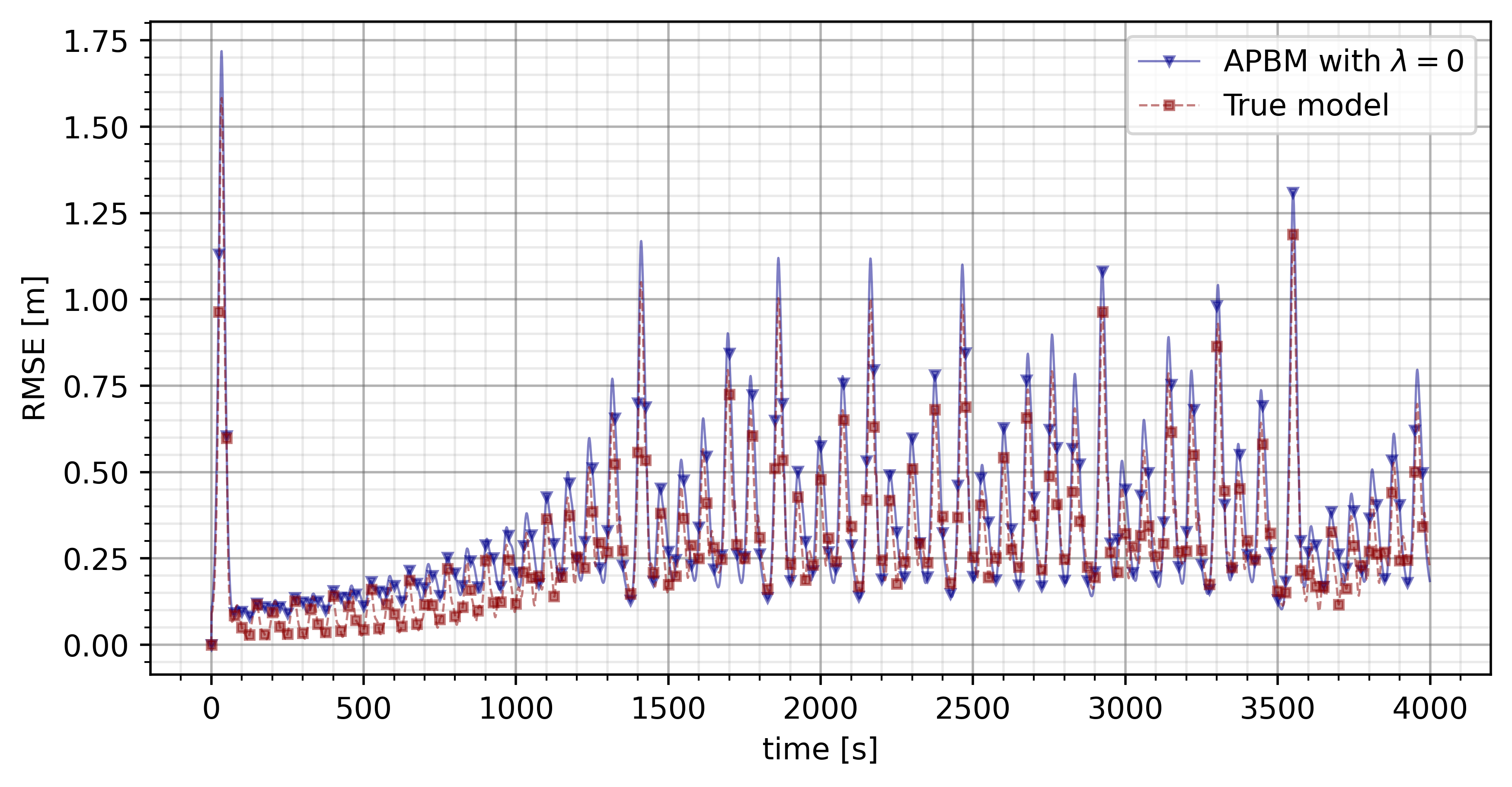}}  
\caption{Result of training on 100 Monte Carlo simulations. Average RMSE for the true model (blue triangles) is compared to the estimations of the APBM approach (orange squares). Note that the APBM is performing worse than the true model in the first $10^3$ seconds but the performance is approaching the true model performance as the parameters of the APBM are being trained. RMSE across different Monte Carlo simulations are similar due to the usage of same the $\rho$, $\beta$, and $\sigma$ parameters throughout the simulations.}
\label{fig:lorenz_rmse}
\end{figure}

The experiment results are summarized in Figures~\ref{fig:neuron_FFTT} and~\ref{fig:lorenz_rmse}. 
Figure~\ref{fig:neuron_FFTT} depicts the Lorenz Attractor ground-truth and the estimation using the APBM with $\lambda = 0$ and the estimation using the true model. 
We considered only $\lambda = 0$ for the APBM since in this case we completely replaced the ODE corresponding to the state $x_1$, and therefore any $\bar{\cb{\theta}}$ satisfying $g\left(f(\bx_{k}), \bx_{k}; \cb{\theta}=\bar{\cb{\theta}}\right) =  f(\bx_{k})$ would be the true ODE. Analyzing Fig.~\ref{fig:neuron_FFTT} we can observe that, even under noisy observations, both filtering processes were able to approximate the chaotic system reasonably well. Although at first sight it seems that the APBM better approximates the Attractor's ground truth, the RMSE over MC realizations depicted in Fig.~\ref{fig:lorenz_rmse} shows a small decrease in performance of APBM with respect to the true model. 



\subsection{Application to target tracking} \label{subsec:target}


State estimation is widely used in many navigation and tracking related applications \cite{dunik2020state}, which would benefit from the proposed data augmentation models.
To test the discussed approach, we consider a two-dimensional target tracking application with additive constant velocity~\cite{closas2011particle} and sinusoidal~\cite{arasaratnam2009cubature} terms. Measurements from two collocated sensors measuring received signal strength (RSS) and bearings were considered:
\begin{equation}\label{eq:target_meas_eq}
{\mathbf y}_k = \left( \begin{array}{c}
                         10 \log_{10} \left( \frac{\Psi_0}{\parallel {\mathbf p}_0 - {\mathbf p}_k \parallel^q} \right) \\
                         \angle ({\mathbf p}_0, {\mathbf p}_k )
                       \end{array}
\right) + {\mathbf n}_k ~,
\end{equation}
\noindent with ${\mathbf p}_0$ being the position of the sensors, ${\mathbf p}_k$ the unknown position of the target,
$10 \log_{10} \left( \Psi_0 \right) = 30$ dBm, $q=2.2$ the path loss
exponent, $\angle ({\mathbf p}_0, {\mathbf p}_k )$ denoting the
angle between locations ${\mathbf p}_0$ and ${\mathbf p}_k$ in radians, and
${\mathbf n}_k \sim {\mathcal N} \left( {\mathbf 0}, \textrm{diag}(1 , 0.1) \right)$ the measurement noise. Sensors were located at the origin of the coordinate system, ${\mathbf p}_0=(0,0)^\top$.

The dynamics of the target were simulated from
\begin{align}\label{eq:target_state_eq}
  {\mathbf x}_k &= (\bF +\bG_{k-1}) {\mathbf x}_{k-1} + \bM {\mathbf u}_{k-1} ~, \\
  \Omega_k &= \Omega_{k-1} + v_{k-1} \nonumber
\end{align}
with 
$$
\bF = \left(\begin{array}{cccc}
                          1 & T_s & 0 & 0 \\
                          0 & 1 & 0 & 0 \\
                          0 & 0 & 1 & T_s \\
                          0 & 0 & 0 & 1
                        \end{array}
  \right), 
  $$
  $$
  \, \bG_{k-1} = \left(\begin{array}{cccc}
                          0 & \frac{\sin\Omega_{k-1} T_s}{T_s} & 0 & -\frac{1 - \cos \Omega_{k-1}T_s}{\Omega_{k-1}} \\
                          0 & \cos \Omega_{k-1}T_s & 0 & -\sin \Omega_{k-1}T_s \\
                          0 & \frac{1 - \cos \Omega_{k-1}T_s}{\Omega_{k-1}} & 0 & \frac{\sin \Omega_{k-1}T_s}{\Omega_{k-1}} \\
                          0 & \sin \Omega_{k-1}T_s & 0 &  \cos \Omega_{k-1}T_s
                        \end{array}
  \right)
  $$
\noindent ${\mathbf x}_k = (x_k, \dot{x}_k, y_k, \dot{y}_k)^\top$
being a state vector, composed of the two-dimensional position
(${\mathbf p}_k = (x_k, y_k)^\top$) and velocity of the target ($\dot{\mathbf p}_k = (\dot{x}_k, \dot{y}_k)^\top$), respectively, and $\Omega_k$ being an angle state. In
(\ref{eq:target_state_eq}), $T_s=1$s is the sampling period and
${\mathbf u}_k\sim {\mathcal N} \left( {\mathbf 0}, 0.1 \cdot
{\mathbf I} \right)$ is the process noise for vector of position and velocities, while $v_k\sim\calN(0, 0.1)$ is the process noise for the angle $\Omega_k$ . The true trajectory was initialized
at
\begin{equation}
{\mathbf x}_0 = (100, 100, 0, 0)^\top
\end{equation}
\noindent and the estimated $\hat{\mathbf x}_0$ was drawn from a Gaussian distribution with mean ${\mathbf x}_0$ and covariance $\textrm{diag} \left(0.1, 0.1, 0.01, 0.01 \right)$.

With this setup, the trajectory described by a target moving for
$T=500$ seconds was generated. Results were averaged over $100$
independent Monte Carlo runs and trajectories, and thus the results
are trajectory-independent. 

To test the capability of APBM models to model unknown parcels of the model, we augment a constant velocity model with a neural network. More precisely we consider:
\begin{equation}
g(f(\bx_{k-1}), \bx_{k-1}; \cb{\theta}) = \bF\bx_{k-1} + \gamma(\bx_{k-1}; \cb{\theta})    
\end{equation}
where $\gamma(\cdot)$ is a neural network with one hidden layer with 5 neurons and ReLu activation function, an output layer with 4 neurons and linear activation, and parameterized by $\cb{\theta}$, leading to a total of 49 parameters including bias terms. As showed in~\eqref{eq:target_state_eq} the term $\bF\bx_{k-1}$ is a simple constant velocity model.


\begin{figure}[htb]
\centerline{\includegraphics[trim={0 0 0 0.1cm},clip,width=\columnwidth]{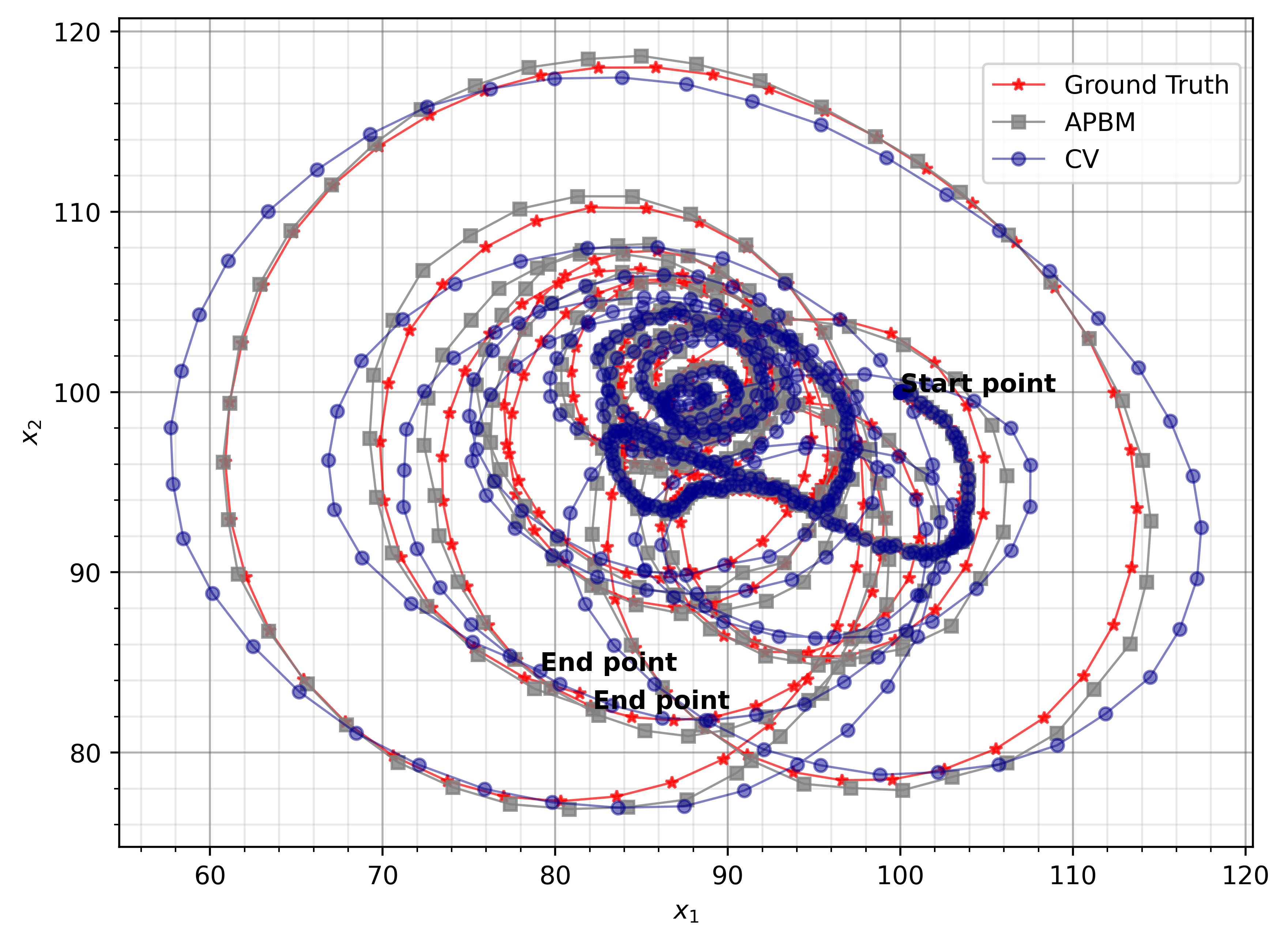}}  
\caption{Comparison of APBM and CV (constant velocity model) for a single simulation. APBM with $\lambda=10$ (gray squares) tracks the ground truth path (red stars) with a smaller error compared to constant velocity model (CV) (blue squares).}
\label{fig:apbm_vs_constant}
\end{figure}

\begin{figure}[htb]
\centerline{\includegraphics[trim={0 0 0 0.1cm},clip,width=\columnwidth]{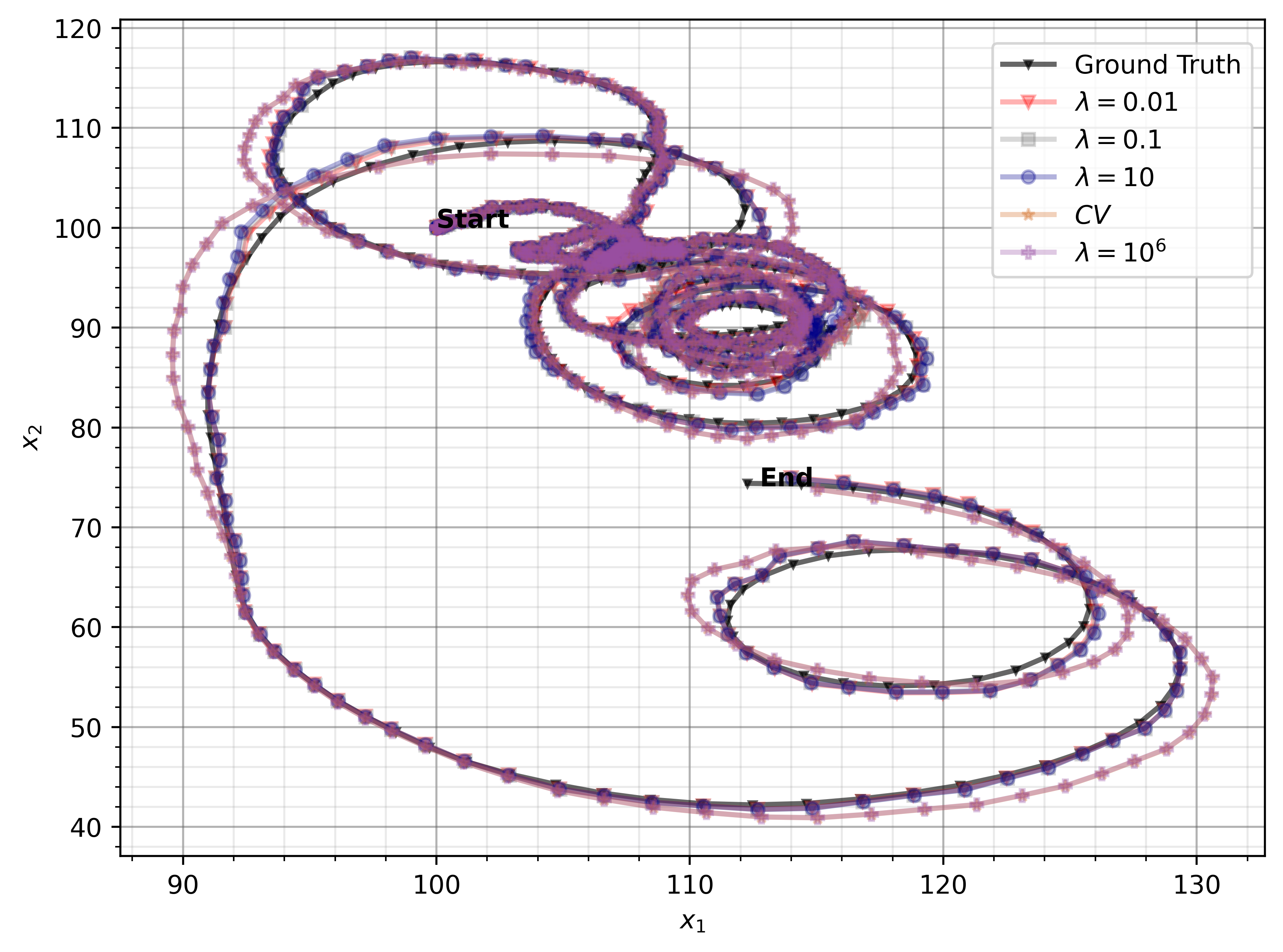}}  
\caption{Result of a single tracking simulation emphasizing the contribution of $\lambda$ on the tracking performance. Black line is the ground truth. $\lambda=0.01$ (red triangles), $\lambda=0.1$ (gray squares), and $\lambda=10$ (blue circles) are in a range for $\lambda$ that results in low RMSE tracking. $\lambda=10^6$ (purple pluses) and CV (orange squares) are overlapping.}
\label{fig:single_tracking_lambda_comp}
\end{figure}

\begin{figure}[htb]
\centerline{\includegraphics[trim={0 0 0 0.1cm},clip,width=\columnwidth]{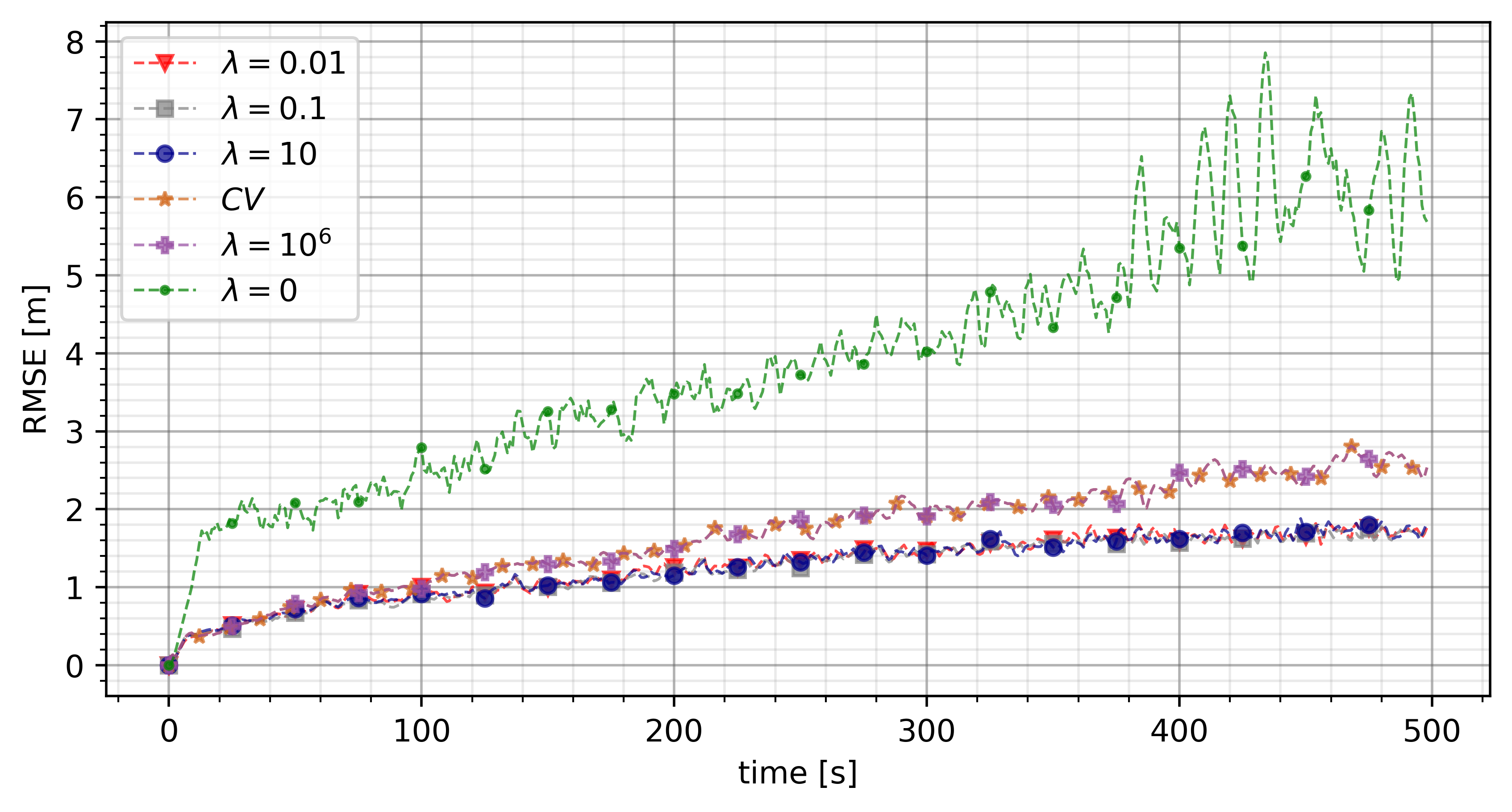}}  
\caption{Result of training on 100 Monte Carlo simulations. Average RMSE for Monte Carlo simulations using different $\lambda$ values are reported. Dashed lines are representing estimated values. RMSE values for the constant velocity model (orange stars) and APBM regularized with $\lambda=10^6$ (purple plus) are overlapping. }
\label{fig:small_large_lambda_vs_constant}
\end{figure}

Figure~\ref{fig:apbm_vs_constant} presents one realization of the experiment where the ground-truth and estates estimates with APBM (with $\lambda = 10$) and constant velocity (CV) model are depicted. 
It can be observed that the APBM approach can track with a smaller error due to the flexibility acquired with the NN component. As expected, the improvement of the APBM with respect to the CV model can be especially noticed in sharper curves. Figure~\ref{fig:single_tracking_lambda_comp} depicts another realization now including APBM with different $\lambda$ values. While the performance of $\lambda \in [0.01, \ldots, 10]$ can be hardily distinguished, the estimations obtained using the CV model and APBM with $\lambda=10^6$ overlap and are clearly worse. This behavior can be more consistently demonstrated in Fig.~\eqref{fig:small_large_lambda_vs_constant} where the evolution of the RMSE computed over MC realizations for CV and APBM, with different values of $\lambda$, models are shown. 
%

Further analyzing Fig.~\ref{fig:small_large_lambda_vs_constant}, we notice that $\lambda=0.01$, $\lambda=0.1$, and $\lambda=10$ yielded to similar RMSE reaching RSMEs under 2m for $t=500$s, that $\lambda=0$ led to the worse result with RMSE above 5m, and, as expected, the RMSE for the APBM with very high $\lambda$ and the CV models are almost identical, with RMS just under 3m. These results indicate that including the constraints over the neural network parameters may lead to improved performance over the unconstrained version and that $\lambda\to\infty$ leads to the CV model as expected.

\begin{figure}[htb]
\centerline{\includegraphics[trim={0 0 0 0.1cm},clip,width=\columnwidth]{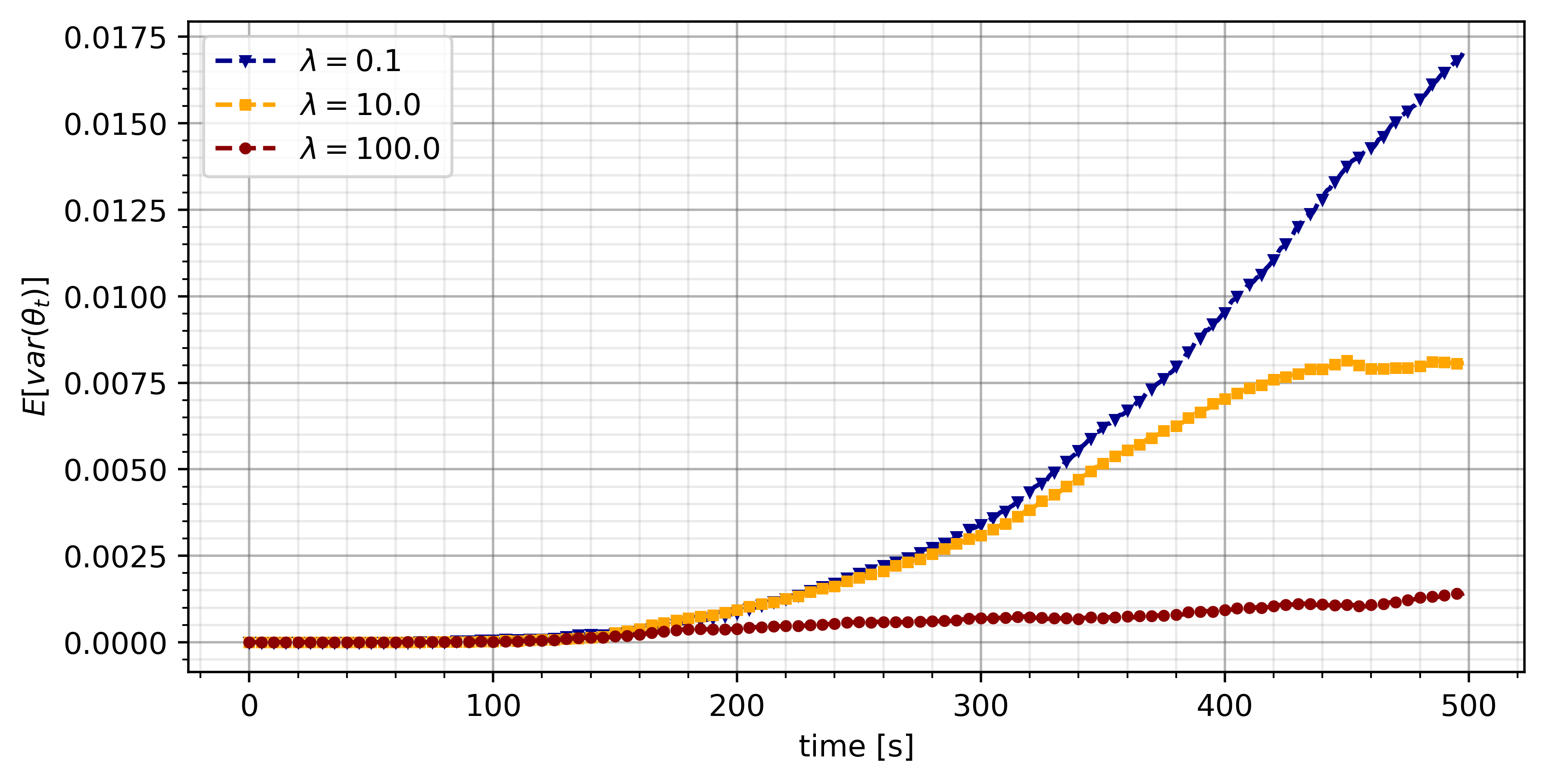}}
\caption{Mean of variances of weights across all 100 Monte Carlo simulations. Eq. \ref{eq:var_formula} is used for all lines. The x-axis demonstrates the time and the y-axis demonstrates the mean of variance of all weights at a time step. Weights are initialized as zeros, and the evolution of the weights in time for different $\lambda$ values is reported. Note that weights trained with a greater regularization term $\lambda$ have a smaller variance.}
\label{fig:var_mean_comparison}
\end{figure}

Finally, we observed the effect of $\lambda$ in the evolution of the neural network parameters. We utilized the Eq. \eqref{eq:var_formula} to calculate the mean of the variance of parameters at a step $k$ for all simulations. Figure \ref{fig:var_mean_comparison} demonstrates that increasing the $\lambda$ forced the parameters to remain close to 0 and we can regularize the parameters by adjusting $\lambda$ accordingly. 

\section{Conclusion}\label{sec:conc}

In this paper, we present a systematic neural augmentation approach centered on physical-based models. The motivation for such an approach is rooted in the desire to produce meaningful and interpretable results while making simplistic models more flexible and capable of adapting to different operation scenarios. Furthermore, we presented a simplistic, yet efficient, strategy to control the neural network contribution to the overall model. We showed with simulations that $i)$ the whole of such constraint over the NN parameters is not negligible since it led to better results when compared with the unconstrained version; and $ii)$ that our intuition regarding $\lambda$, discussed in Section~\ref{sec:Control}, was correct.

\bibliographystyle{IEEEbib}
\bibliography{main}

\end{document}